# AUTOMATED CROP FIELD SURVEILLANCE USING COMPUTER VISION

Tejas Atul Khare
School of Electronics and
Communication Engineering
MIT World Peace University
Pune, India
tejas.khare99@gmail.com

Anuradha C. Phadke
School of Electronics and
Communication Engineering
MIT World Peace University
Pune, India
anuradha.phadke@mitwpu.edu.in

*Abstract*— Artificial Intelligence is everywhere today. But unfortunately, Agriculture has not been able to get that much attention from Artificial Intelligence (AI). A lack of automation persists in the agriculture industry. For over many years, farmers and crop field owners have been facing a problem of *trespassing of wild animals* for which no feasible solution has been provided. Installing a fence or barrier like structure is neither feasible nor efficient due to the large areas covered by the fields. Also, if the landowner can afford to build a wall or barrier, government policies for building walls are often very irksome. The paper intends to give a *simple intelligible solution* to the problem with *Automated Crop Field Surveillance using Computer Vision*. The solution will significantly reduce the cost of crops destroyed annually and completely automate the security of the field.

## I. INTRODUCTION

### A. Background

In many countries like the USA, Australia, Canada, India, farmers, and landowners from many centuries are struggling to tackle the problem of wild animal trespassing in their fields. Framers have to stay up all night to manually scare the animals away. Heavyweight animal species like elephants, bears, and pigs often trespass for grazing and destroy the crops in the process. Smaller wild animals like monkeys, oxen, horses, and deer also have a significant contribution to the same. Using fences to protect the crops will not be feasible because of the vast area covered by the farm. The amount of capital loss is enormous and has to be eliminated.

### B. Literature Survey

It is estimated that 30 – 40 percent of crops are destroyed annually due to attacks by wild animals in India. In some districts of Odisha and Kerala (India), elephants destroy up to 60 percent of crops the claims for which were not payable by insurance companies. Losses of grain in Canada have become so serious that major efforts have been made to alleviate them. Devices of various kinds to frighten the animals, and spreading grain near the marshes to lure the animals from the unharvested grain fields, are two methods that have been used. Both are cumbersome and expensive, and they are not always effective. Taking the point of fencing drawbacks further, Puja Mondal [15] illustrates the costs incurred while installing a fence. The price of the fence is Rs. 6000 per 365 meters on average and hence that comes to Rs. 16.4 per meter. Hence, assuming a one-acre area, the cost for only purchasing a fence will be Rs. 1,70,761. Furthermore, the cost of labour and other necessary prerequisites of installing a fence will add up to Rs. 5,50,920 which is $7540 for an acre. Heavyweight animals like a pig or an elephant will trespass the fence, hence the farmer would suffer from a huge loss of capital.

In the United States, wildlife damage to agricultural resources is significant. Damages to crops because of wildlife on an average are $961 per hectare which is more than an average farmer earns from a hectare [1]. Field crop losses to wildlife totalled $619 million and losses of vegetables, fruits, and nuts totalled $146 million.
In the United States, nearly 3800 acres of loss were reported by farmers in 2009[2]. The statistics [2] suggest that crops lost in the country due to White-tailed deer (94%), Wild Hogs (61%), Coyotes (33%), Raccoons (30%), Armadillos (19%), Rabbits (10%) are significantly large. In addition to the animals, birds are also an important factor responsible for the destruction of the crops. Small plants and crops like, sunflowers, wheat, rice, etc. are severely affected by birds. In 2012, the loss of fruits exceeded $41 million [2] and as sunflowers and rice are the main crops, farmers face a huge loss of around $5 – 13 million.
Instead of using traditional CCTV (Closed Circuit Television) cameras for monitoring the fields manually, Computer Vision technique can be implemented which provides an extension of automation with the same traditional CCTV camera. Computer Vision has emerged as a promising solution to problems concerned with surveillance like outdoor perimeter monitoring system [7]. In addition to the applications in surveillance, Computer Vision has gained significant popularity in industrial applications. Hernan H. Álvarez-Valera et al [8] introduce a solution to industrial problems that involve manual labour. They have proposed an automatic Chestnut Selection system which works on the features like the oval shape for detecting the chestnut product and colour and size for defect detection. Aniruddha et al [13] implemented a Locally-connected Neural Pyramid (LCNP) using the CUDA platform for recognition of objects from disparate classes like a person, car, building, etc. In this work, they achieved speedy training of large datasets and a recognition rate of 85.62% for the testing samples. Vrushali Pagire et al [14] developed an FPGA based system for detecting a moving object for surveillance application using background subtraction algorithm and dynamic thresholding. Vikas Bavane et al [12] propose to prevent animal trespassing by using Raspberry Pi and installing Passive Infrared (PIR) sensors for motion detection. The system checks the authority of the detected object with the help of an RFID tag. If the detection is of an unauthorized object, the buzzer makes a sound after confirming the presence of an animal.

The purpose of this literature survey was to address the problems faced by farm owners and farmers. The problem is significant because of the huge amount of capital loss and hence, providing some related research on the problem.

## II. HYPOTHESIS

The hypothesis is based on Computer Vision, primarily on Object Detection. The Proof of Concept revolves around Object Detection and taking action based on the detections obtained.

Research questions:
1. Can the idea be implemented in the real world?
2. Is the solution feasible?
3. How efficiently does it solve the problem?

## III. PROPOSED METHOD

The paper intends to solve the problem by placing long-range cameras at the corners of the field or land with considering the maximum field of view of the camera. Speakers with an equal amount of distance between them are placed inside the field. The positions of the camera and the speakers are static. The video from the camera is captured and broken down into frames which are then sent sequentially to the object detection module. The time interval between the successive frames goes up to 1.5 seconds depending on the hardware acceleration. For this proposed solution, a 1.5 seconds lag would not significantly affect the efficiency of preventing the animals to enter the field. When an animal enters the field of view of the camera Fig. 5, the object detection module would first classify if it is a potential threat to the farm. If the animal is classified as a threat, the distances between the animal and the speakers inside the farm which are in the field of view Fig. 5 is calculated. To decide which speaker should make a sound, the speaker closest to the animal is identified by filtering the other speakers with a larger distance.

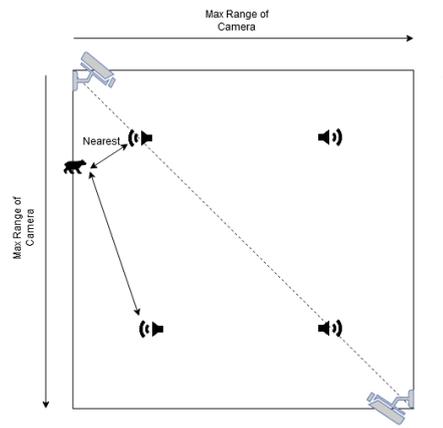

Fig. 1 Structure of the system (Top View)

In Fig. 1, the distance of the animal is calculated between the two speakers which are inside the field of view of the camera Fig. 5. From the top view, it can be explicitly seen that the speaker nearer to the animal will be played. The number of speakers to be installed depends upon the maximum range of the camera. The cameras would cover their two adjacent edges effectively if it has at most 90° field of view. Ensuring a maximum field of view would significantly reduce the number of cameras and speakers needed for the same field.

The object detection is carried out by a pretrained model – YOLOv3 [3]. The model uses the COCO [6] dataset which consists of 80 classes. The classes in the dataset include the animals responsible for damaging the crops – 'horse', 'sheep', 'cow', 'elephant', 'bear', and many irrelevant classes that are not required for the model to train on. Hence, for this solution, the model is trained for detecting speakers and wild animals like pigs, boars, etc in addition to the classes of animals present in the COCO dataset. The irrelevant classes in the COCO set are discarded. The input images of speakers fed for training are labelled with Labelimg [9]. The annotation files are obtained after labelling and then fed to the model with the corresponding image for training [4]. After the completion of training, a weights file is obtained which can be loaded to the neural network using OpenCV [5].

*Note:* Fig. 2 a) and b) depicts the real-time working of the system. The blue bounding boxes for the animals and birds are obtained by the object detection module. The red bounding boxes are for the speakers. The horn speakers are manually placed with the help of MS paint software in the original input image to show the working of the system.

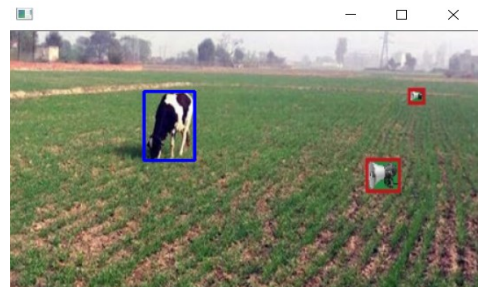

(a)

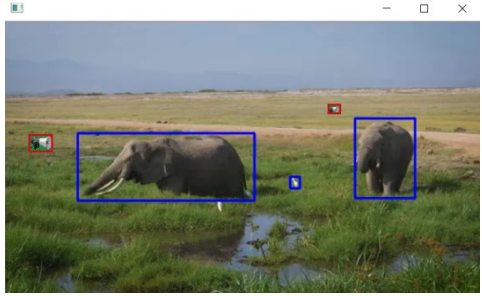
(b)
Fig. 2 Working of System (a) Animal detected-cattle; (b)Animal detected-Elephant, bird

Fig. 3 represents the flow of the system. The cameras are connected to a video multiplexer and the video feeds are passed to the object detection model. The object detection model detects speakers continuously and hence, each speaker has a unique ID which is a simple integer. Once an animal is detected, the nearest speaker to the animal is identified by calculating the distance of the animal from each speaker for cost-cutting, instead of playing all the speakers. After the speaker that is closest to the animal is identified, the ID of that speaker is chosen to send the output audio. This speaker selection is done with the help of python module sounddevice [16] using a query that gets the list of connected speakers and passing the ID to the output inbuilt function. Thus, only the speaker nearest to the animal is played (In Fig. 3, speaker 3 is selected).

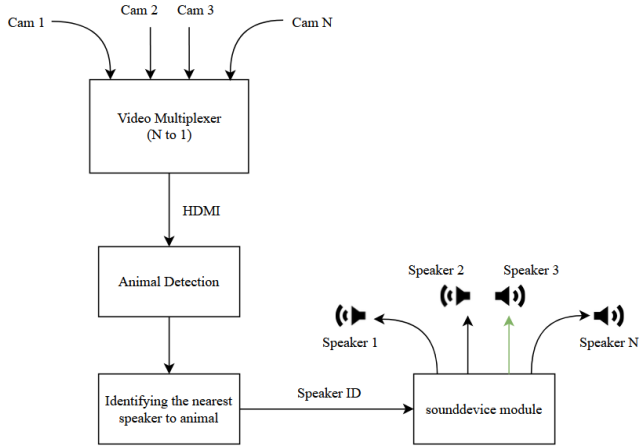
Fig. 3 Block Diagram of System

## IV. THEORY

*Computer Vision* is a subset of Artificial Intelligence where a computer is trained to interpret and understand the visual world or the world as 'we see it'. In other words, it enables the computer to gain a high-level understanding of the visuals just like humans do. The training is established by feeding the computer a large number of images and videos from cameras. These images are then passed to deep learning models for feature extraction. After the features are extracted, the computer is ready to understand the visual world. This solution uses YOLO for object detection.

### A. YOLO

You Only Look Once (YOLO) is a classification network that has an architecture of a Fully Convolutional Neural Network (FCNN) which passes an input image of n x n size only once through the FCNN unlike the other classification networks like Fast RCNN. The size of the output image is m x m which is set in the configuration file of the model. The input image is split into s x s grid and multiple bounding boxes are predicted for a single object. The bounding box with the highest probability is then selected and others are discarded by using Non Maximum Suppression [10].

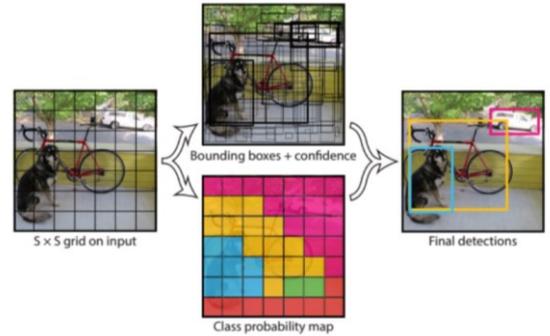
Fig. 4 Working of YOLO

### B. Field of View

The field of view is that part of the world that is visible through the camera at a particular position and orientation in space; objects outside the FOV when the video is recorded are not a part of the video field. It is most often expressed as the angular size of the view cone, as an angle of view. The camera is positioned in such a way that the field of view of the camera covers a maximum area of the field.

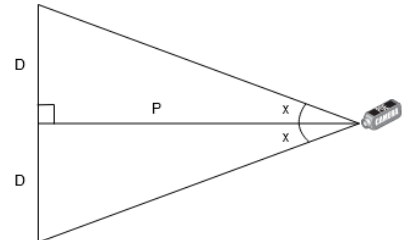
Fig. 5 Calculation of Angle of View

$$\tan x = D/P \quad (1)$$
$$D = P \times \tan x \quad (2)$$
$$2D = 2 \times P \times \tan x \quad (3)$$

2D is the maximum horizontal distance in the field of view of the camera

$$x = \tan^{-1} D/P \quad (4)$$

As 2x is the Angle subtended by the camera,

$$2x = 2 \times \tan^{-1} D/P \quad (5)$$

After the angle of view is calculated from Fig. 5, the exact field of view can be obtained from the camera for placing the speakers.

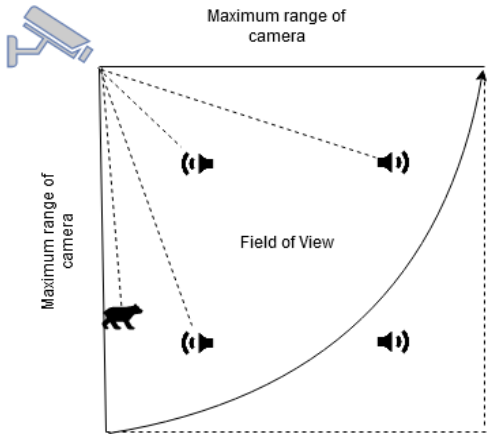

Fig. 6 Field of View of Camera

The animal will be detected when it is in the field of view of the camera Fig. 6. After the detection of the animal, the distance between the speakers and the camera is calculated.

C. *Calculating the Shortest Distance between two objects in an image*

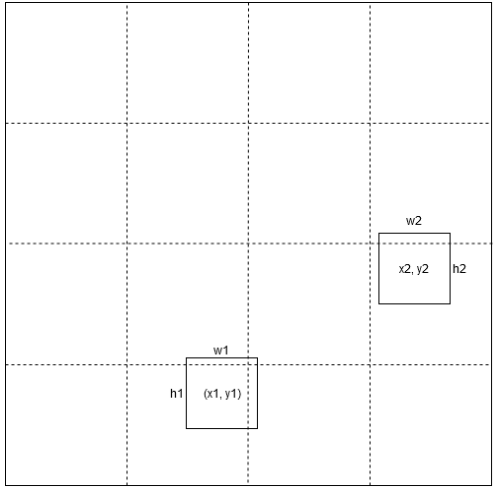

Fig. 7 Detected objects in an image

Fig. 7 illustrates the drawing of bounding boxes after Non Maximum Suppression [12]. After the detection of objects, and getting the coordinates of bounding boxes, the distance between the object is calculated. In Fig. 7,

$(x1, y1)$ and $(x2, y2)$ are coordinates of the centre of the bounding boxes respectively,

$(w1, h1)$ and $(w2, h2)$ are the width and height of the bounding boxes respectively

$x1, y1, x2, y2, w1, h1, w2$ and $h2$ are the values concerning the pixels which are dependent on the image resolution e.g. $416 \times 416$.

To calculate the distance between the objects, the following algorithm is used-

1. Calculate Euclidian Distance

$$d = \sqrt{(x1 - x2)^2 + (y1 - y2)^2} \qquad (6)$$

Equation (6) gives the formula for computing the distance of the objects (detected animal and speakers) between two pixels.

2. Calculate actual distance between the objects

Using the Focal Length, the maximum Range, and the Pixel Width (sensor) of the camera, the IFOV (Instantaneous Field of View is calculated by using equation (7) -

IFOV – Angle subtended by each pixel of the camera in the field

$$\text{IFOV} = \frac{\text{Pixel Width}}{\text{Focal Length}} \qquad (7)$$

$$\text{actual distance covered by a single pixel} = \text{Range(m)} \times \text{IFOV} \qquad (8)$$

Hence, actual distance (AD) between objects from equation (8) is given by -

$$=> \text{AD(m)} = \text{Range(m)} \times \text{IFOV} \times d \qquad (9)$$

In (8) and (9), Range(m) of a camera is the distance at which detail can be captured.

V. EXPERIMENTAL TESTING OF THE ALGORITHM FOR DISTANCE CALCULATION

*The purpose of this experiment is to validate the accuracy of the algorithm to work on the objects that are smaller than the size of the objects interpreted in this solution that are animals and a speaker and also for the algorithm to work with any detectable objects*, hence in this experiment; pre-trained weight file of COCO dataset [6] was used with YOLO object detection technique. The dataset contains 80 classes which include the objects used for the experiment - 'bottle', 'backpack', 'cup'. This is not the most robust method to test the algorithm for this particular solution, but it significantly produces results that can be used to validate the proof of concept. The experiment was conducted in different backgrounds and light conditions and was set up to validate the performance of the algorithm in the unfavorable conditions when the camera is not placed at a height greater than of the objects. One case where the camera is placed at a height greater than the object is also considered Fig. 9 (b). Fig. 8 and Fig. 9 illustrates the experimental setup under the conditions mentioned.

*Note*: The images are taken locally by the author, no copyrights were disregarded.

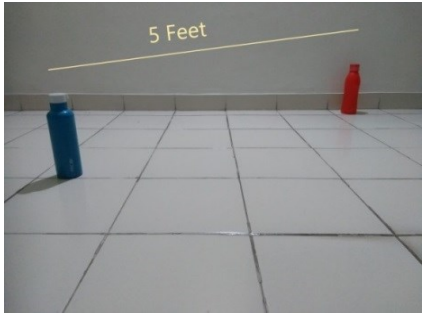
(a) 5 Feet

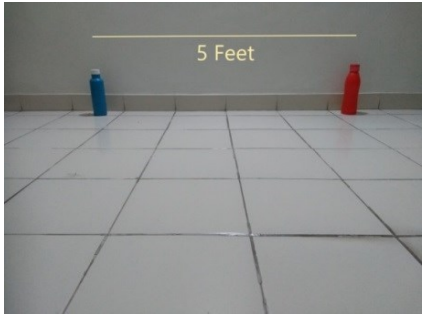
(b) 5 feet perpendicular to the camera

Fig. 8 Experimental Setup of the three cases

## A. Results

The following results were observed by using a similar setup as Fig. 8. The bounding boxes are obtained by the object detection model. The floating-point numbers above an object in the image in Fig. 8 indicate the distance between that object and its neighbouring object.

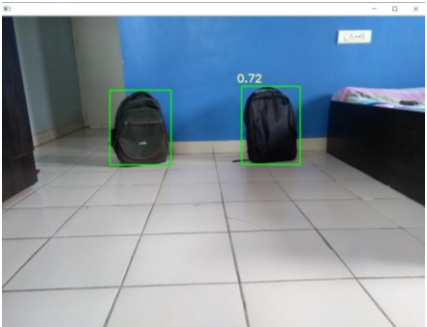
(a) Backpack 0.91 meters apart

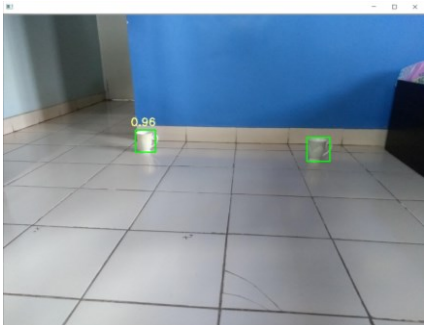
(b) Cups 0.91 meters apart

Fig. 9 Experimental Results

TABLE I.  OUTPUT OF TEST CASES

| Test Case | Obtained Distance | Actual Distance | % Error |
|---|---|---|---|
| Bottles | 1.79 | 1.52 | +17.7 |
| Bottles Perpendicular | 1.54 | 1.52 | +1.3 |
| Backpack Perpendicular | 0.72 | 0.91 | -20.8 |
| Cups | 0.96 | 0.91 | +5.5 |

Fig. 10 Output of Experimentation

## B. Observations

The experimentation was conducted on 8 different objects 'bottles', 'backpack', 'cup', 'sports ball', 'bowl', 'banana', 'carrot', 'mouse', and 'spoon', each with 2 different cases of placing the objects.
1. Perpendicular to the camera as shown in Fig. 8 b), Fig. 9 a)
2. Angles other than 90° Fig. 8 a), Fig. 9 b)

The results lucidly suggest that all the objects were detected under different light and background conditions. Also, it was observed that the % error in the algorithm increases when the objects are at an angle with respect to the camera. Moreover, if the objects are relatively closer to the camera, an error towards the negative side is seen. The relation between the sizes of the objects is also seen to have an impact on efficiency. The ratio of the size of objects to the distance between them is seen to have an impact on the accuracy as it is expected. When the distance between them is increased, the error reduces.

Furthermore, the error is insignificant for the perpendicular case. Hence, on average, a +8.16% and -20.8% error is seen in the algorithm.

## VI. CONCLUSION

In this paper, a novel solution was introduced to automate the crop field surveillance by using computer vision by presenting a literature survey of the importance of this solution. Furthermore, an algorithm was introduced to calculate the shortest distance between objects in an image. To validate the algorithm, a similar work of the author was introduced to calculate the distance between two people for social distancing.

By the observations, it can be concluded that as the ratio increases the efficiency decreases. Furthermore, the size of the object with respect to the camera contributes to the accuracy equally. The error recorded is +8.16% and -20.8%. But in the case of the solution, the animal and the speaker are perpendicular to the camera and would be at a significant distance from the camera. Hence, the probability of the solution to be feasible and efficient is significantly high. Considering these errors, if an animal enters the field, there would be a delay for the speakers to make a sound but the delay would not contribute enormously. For instance, consider a bear that has an average walking speed of 1.7 m/s [11]. If the bear is at 1.52 m from the speaker but the calculated distance shows 1.79 m, the delay of the speaker's initiation of sound will be 0.15 sec which is insignificant. This solution could significantly reduce the crops lost annually, hence reducing the number of efforts and capital required to control such events. Even if the system works at 40% efficiency, according to the National Agricultural Statistics Service (NASS) in the United States of America [1], we could save $377.6 million.

## VII. Future Work

The solution needs to be validated in different circumstances like in the night, or where there is not a significant amount of light. Furthermore, the model can be trained for detecting animals and speakers at night for better performance. The hard wire nature of the system limits it to a definite number of cameras (N<=32) because of the availability of N to 1 video multiplexers as shown in Fig. 3.

## Acknowledgment

This paper would not have been possible without the open source object detection model YOLO. The authors would like to thank the reviewers for taking out their precious time for reviewing this paper. This paper is dedicated to the hardworking farmers to help them in any possible way.